\documentclass[10pt,twocolumn,letterpaper]{article}

\usepackage{cvpr}
\usepackage{times}
\usepackage{epsfig}
\usepackage{graphicx}
\usepackage{amsmath}
\usepackage{amssymb}
\usepackage{algorithm}
\usepackage[noend]{algpseudocode}
\usepackage{float}
\usepackage{subfig}
\newcommand{\refsec}[1]{Sec\onedot~\ref{#1}}
\newcommand{\reffig}[1]{Fig\onedot~\ref{#1}}
\newcommand{\refeq}[1]{Eq\onedot~\ref{#1}}
\newcommand{\reftab}[1]{Table~\ref{#1}}

\usepackage[pagebackref=true,breaklinks=true,letterpaper=true,colorlinks,bookmarks=false]{hyperref}

\cvprfinalcopy 

\newcommand{\omitme}[1]{}

\def\cvprPaperID{3218} 

\ifcvprfinal\fi
\begin{document}

\title{Deep Spatio-Temporal Random Fields for Efficient Video Segmentation}

\author{Siddhartha Chandra$^{1}$ \hspace{20mm} Camille Couprie$^{2}$ \hspace{20mm} Iasonas Kokkinos$^{2}$\\
\hspace{-10mm} {\tt \small siddhartha.chandra@inria.fr} \hspace{11mm} {\tt \small coupriec@fb.com} \hspace{23mm} {\tt \small iasonask@fb.com}\\
\vspace{4mm}
$^1$ INRIA GALEN, Ecole CentraleSup\'elec Paris \hfill $^2$ Facebook AI Research, Paris\\}
{}

\maketitle

\begin{abstract}

In this work we introduce a time- and memory-efficient method for structured prediction that couples neuron decisions across both space at time. We show that we are able to perform exact and efficient inference on a densely-connected spatio-temporal graph by capitalizing on recent advances on deep Gaussian Conditional Random Fields (GCRFs).
Our method, called VideoGCRF is (a) efficient, (b) has a unique global minimum,
and (c) can be trained end-to-end alongside contemporary deep networks for video understanding.
We experiment with multiple connectivity patterns in the temporal domain, and present empirical improvements over strong baselines on the tasks of both semantic and instance segmentation of videos. 
 Our implementation is based on the \emph{Caffe2} framework and will
be available at \url{https://github.com/siddharthachandra/gcrf-v3.0}.
\end{abstract}

\section{Introduction}

Video understanding remains largely unsolved despite  significant improvements  in image understanding over the past few years. 
The accuracy of current image classification and semantic segmentation models is not yet matched in  action recognition and video segmentation, to some extent due to the lack of  large-scale benchmarks, but also due to the complexity introduced by  the time variable. Combined  with the increase in memory and computation demands, video understanding poses additional challenges that call for novel methods. 

Our objective in this work is to couple the decisions taken by a neural network in time, in a manner that allows information to flow across frames and thereby result in decisions that are consistent both spatially and temporally. Towards this goal we pursue a structured prediction approach, where the structure of the output space is exploited in order to train classifiers of higher accuracy. 
For this we introduce VideoGCRF, an extension into video segmentation of the Deep Gaussian Random Field (DGRF) technique recently proposed for single-frame structured prediction in \cite{gcrf,gcrf2}. 

\begin{figure}[t]
    \centering
    \includegraphics[width=\linewidth]{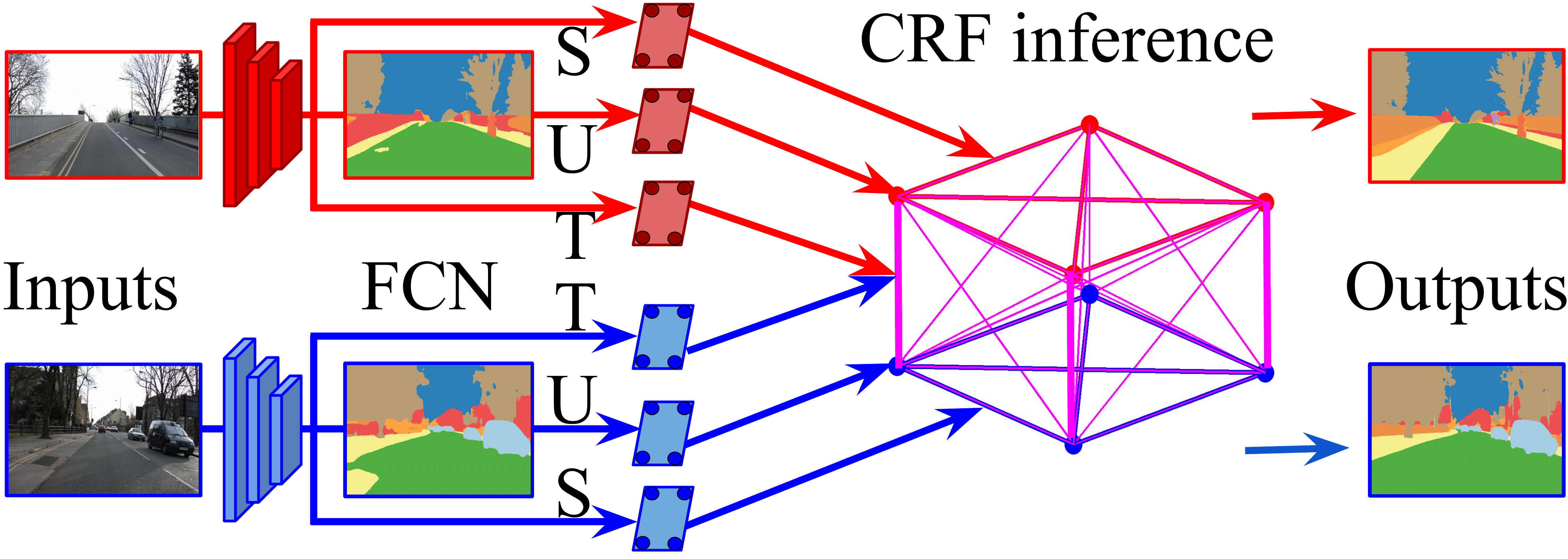}
    \caption{Overview of our VideoGCRF approach: we jointly segment multiple images by passing them firstly through a fully convolutional network to obtain per-pixel class scores (`unary' terms U), alongside with spatial (S) and temporal (T) embeddings. We couple predictions at different spatial and temporal positions in terms of the inner product of their respective embeddings, shown here as arrows pointing to a graph edge. The final prediction is obtained by solving a linear system; this can eliminate spurious responses, e.g. on the left pavement, by diffusing the per-pixel node scores over the whole spatio-temporal graph. The CRF and CNN architecture is jointly trained end-to-end, while CRF inference is exact and particularly efficient. 
    }
    \label{fig:teaser}
\end{figure}

We show that our algorithm can be used for a variety of video segmentation tasks: semantic segmentation (CamVid dataset), instance tracking (DAVIS dataset), and a combination of instance segmentation with Mask-RCNN-style object detection, customized in particular for the person class (DAVIS Person dataset). 


Our work  inherits all favorable properties of the DGRF method: in particular, our method has the advantage of delivering (a) exact inference results through the solution of a linear system, rather than relying on approximate mean-field inference, as \cite{densecrf,fso}, (b) allowing for exact computation of the gradient during back-propagation, thereby alleviating the need for the memory-demanding back-propagation-through-time used in \cite{crfrnn}
(c) making it possible to use non-parametric terms for the pairwise term, rather than confining ourselves to pairwise terms of a predetermined form, as \cite{densecrf,fso}, and (d) facilitating inference on both densely- and  sparsely-connected graphs, as well as facilitating blends of both graph topologies. 

Within the literature on spatio-temporal structured prediction,
the work that is closest in spirit to ours is the work of \cite{fso} on Feature Space Optimization. Even though our works share several conceptual similarities, our method is entirely different at the technical level. In our case spatio-temporal inference is implemented as a structured, `lateral connection' layer that is trained jointly with the feed-forward CNNs, while the method of \cite{fso} is applied  at a post-processing stage to refine a classifier's results.

\subsection{Previous work}
Structured prediction is commonly used by semantic segmentation algorithms  ~\cite{gcrf,gcrf2,deeplab1,deeplab2,schwing,Vemulapalli_2016_CVPR,vu2015context,crfrnn} to capture spatial constraints within an image frame. These approaches may be extended naively to videos, by making predictions individually for each frame. However, in doing so, we ignore the temporal context, thereby ignoring the tendency of consecutive video frames to be similar to each other. To address this shortcoming, a number of deep learning methods employ some kind of structured prediction strategy to ensure temporal coherence in the predictions. Initial attempts to capture spatio-temporal context  involved designing deep learning architectures \cite{karpathy1} that implicitly learn interactions between consecutive image frames. A number of subsequent approaches used Recurrent 
Neural Networks (RNNs) \cite{adi2017sequence,donahue} to capture interdependencies between the image frames. Other approaches have exploited optical flow computed from state of the art approaches \cite{flownet2} as additional input to the network \cite{videocnn,fusionseg}. Finally, \cite{fso}  explicitly capture temporal constraints via pairwise terms over probabilistic graphical models, but operate post-hoc, i.e. are not trained jointly with the underlying network.

In this work, we focus on three problems, namely (i) semantic and (ii) instance video segmentation as well as (iii) semantic instance tracking.
Semantic instance tracking refers to the problem where we are given the ground truth for the first frame of a video, and the goal is to predict these instance masks
on the subsequent video frames. 
The first set of approaches to address this task start with a deep network pretrained for image classification on large datasets such as Imagenet or COCO, and finetune it on the first frame
of the video with labeled ground truth \cite{osvos,onavos}, optionally leveraging a variety of data augmentation regimes \cite{lucid} to increase robustness to scale/pose variation
and occlusion/truncation in the subsequent frames of the video.
The second set of approaches poses this problem as a warping problem \cite{masktrack}, where the goal is to warp the segmentation of the first frame using the images and optical flow as additional inputs \cite{vidprop,lucid,li2017video}. 

A number of approaches have attempted to exploit temporal information to improve over static image segmentation approaches for video segmentation.
Clockwork convnets \cite{ShelhamerRHD16clockwork} were introduced to exploit the persistence of features across time and schedule the processing of some layers at different update rates according to their semantic stability.
Similar feature flow propagation ideas were employed in \cite{fso,ZhuXDYW16DeepFeatureFlow}.
In \cite{NilssonS16SemSegmGatedFlow} segmentations are warped using the flow and spatial transformer networks. 
Rather than using optical flow, the prediction of future segmentations \cite{Jin2016PredictingFutureSegm} may also temporally smooth  results obtained frame-by-frame. Finally, the state-of-the-art on this task \cite{videocnn} improves over PSPnet\cite{pspnet}  by warping the feature maps of a static segmentation CNN to emulate a video segmentation network.

\omitme{
\item Clockwork convnets for video semantic segmentation 
\item Semantic Video Segmentation by Gated Recurrent Flow Propagation
\cite{NilssonS16SemSegmGatedFlow}  
\item \cite{Jin2016PredictingFutureSegm} uses prediction of future segmentations to temporally smooth frame by
  frame predictions (PEARL). 
\item Deep feature flow \cite{ZhuXDYW16DeepFeatureFlow}: runs a CNN only on sparse key frames and propagates their feature maps to other frames via optical flow. Improvements in speed but not accuracy.
\item Improvement over PSPnet performances using optical flow warping : the work of \cite{videocnn} employs warping on the feature maps of a static segmentation CNN to emulate a video segmentation network.
}


\begin{figure*}[h!]
\begin{center}
\includegraphics[width=0.95\linewidth]{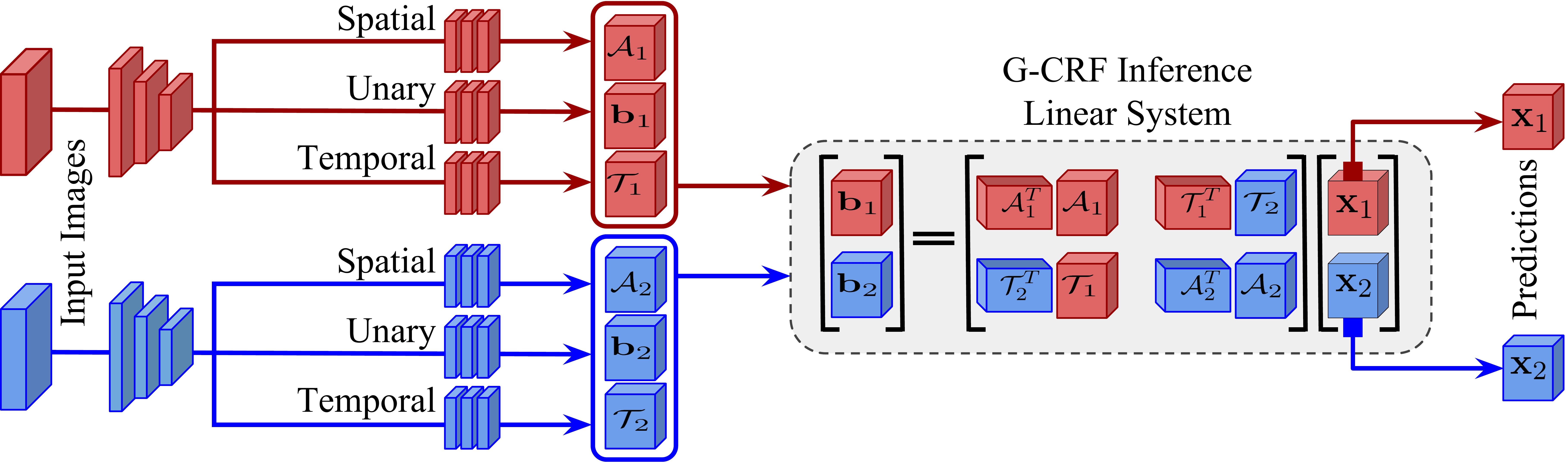}
\end{center}
\caption{VideoGCRF schematic for $2$ video frames. Our network takes in $2$ input images, and delivers the per frame unaries $\mathbf{b}_1,\mathbf{b}_2$, spatial embeddings $\mathcal{A}_1,\mathcal{A}_2$, and temporal embeddings $\mathcal{T}_1,\mathcal{T}_2$ in the feed-forward mode. Our VideoGCRF module collects these and solves the inference problem in \refeq{eqn:stgcrf} to recover predictions $\mathbf{x}_1,\mathbf{x}_2$. During backward pass, the gradients of the predictions are delivered to the VideoGCRF model. It uses these to compute the gradients for the unary terms as well as the spatio-temporal embeddings and back-propagates them through the network.}
\label{fig:temporalGCRF} 
\end{figure*}

\section{VideoGCRF}
In this work we introduce VideoGCRF, extending the Deep Gaussian CRF approach introduced in \cite{gcrf,gcrf2} to operate efficiently for video segmentation.
Introducing a CRF allows us to couple the decisions between sets of variables that should be influencing each other; spatial connections were already explored in \cite{gcrf,gcrf2} and can be understood as propagating information from distinctive image positions (e.g. the face of a person) to more ambiguous regions (e.g. the person's clothes). In this work we also introduce temporal connections to integrate information over time, allowing us for instance to correctly segment frames where the object is not clearly visible by propagating information from different time frames. 

\newcommand{\vid}{\mathcal{V}}
We consider that the input to our system is a video $\vid = \{ I_1, I_2, \ldots , I_V \}$ containing $V$ frames. We denote our network's prediction as $\mathbf{x}_v,~v=1,\ldots,V$, where at any frame the prediction $\mathbf{x}_i \in \mathbb{R}^{PL}$ 
provides a real-valued vector of scores 
for the $L$ classes for each of the $P$ image patches; for brevity, we denote by $N=P \times L$ the number of prediction variables.
The $L$ scores corresponding to a patch can be understood as inputs to a softmax function that yields the label posteriors.

The Gaussian-CRF (or, G-CRF) model defines a joint posterior distribution through a Gaussian multivariate density for a video as:
\begin{equation*}
p(\textbf{x}|\vid) \propto \exp(-\frac{1}{2}\textbf{x}^\top A_{\vid} \textbf{x} + B_{\vid} \textbf{x})\text{,}
\end{equation*}
where $B_{\vid}$, $A_{\vid}$ denote  the `unary' and `pairwise' terms respectively,  with $B_{\vid} \in \mathbb{R}^{NV}$ and $A_{\vid} \in \mathbb{R}^{NV \times NV}$.
In the rest of this work we assume that  $A,B$ depend on the input video
and we omit the conditioning on $\vid$ for  convenience.

What is particular about the G-CRF is that, assuming the matrix of pairwise terms $A$ is positive-definite, the Maximum-A-Posterior (MAP) inference merely amounts to  solving the system of linear equations $A \textbf{x} = B$.
In fact, as in \cite{gcrf}, we can drop the probabilistic formulation and treat the G-CRF as a structured prediction module that is part of a deep network. 
In the forward pass, the unary and the pairwise terms $B$ and $A$, delivered by a feed-forward CNN described in \refsec{cnn} are fed to the G-CRF module which performs
inference to recover the prediction $\textbf{x}$ by solving a system of linear equations given by
\begin{equation}
 ( A + \lambda \mathbf{I} ) \mathbf{x} = B
 \label{eqn:linearSolver},
\end{equation}
where $\lambda$ is a small positive constant added to the diagonal entries of $A$ to make it positive definite.

For the single-frame case ($V=1$)  the iterative conjugate gradient~\cite{conjugategradient} algorithm was used to rapidly solve the resulting system for both sparse \cite{gcrf} and fully connected \cite{gcrf2} graphs; in particular the speed of the resulting inference is in the order of 30ms on the GPU, almost two orders of magnitude faster than the implementation of DenseCRF \cite{densecrf}, while at the same time giving more accurate results.  

Our first contribution in this work consists in designing the structure of the matrix $A_{\vid}$ so that the resulting system solution remains manageable as the number of frames increases. Once we  describe how we structure $A_{\vid}$, we then will turn to learning our network in an end-to-end manner.

\subsection{Spatio-temporal connections}
\label{cnn}


In order to capture the spatio-temporal context, we are interested in capturing two kinds of pairwise interactions: (a) pairwise terms between patches in the same frame and (b) pairwise terms between patches in different frames.

Denoting  the spatial pairwise terms at frame $v$ by $A_{v}$ and the temporal pairwise terms between frames $u,v$ as  $T_{u,v}$ we can rewrite \refeq{eqn:linearSolver} as follows:
\begin{equation}
\begin{bmatrix} A_1 + \lambda \mathbf{I} & T_{1,2} & \cdots & T_{1,V}\\ T_{2,1} & A_2 + \lambda \mathbf{I} & \cdots & T_{2,V}  \\  &  & \vdots  &  \\ T_{V,1} & T_{V,2} & \cdots & A_V + \lambda \mathbf{I} \end{bmatrix} \hspace{-2mm} \left[ \begin{array}{c} \textbf{x}_1 \\ \textbf{x}_2 \\ \vdots \\ \textbf{x}_V \end{array} \right] \hspace{-1mm} = \hspace{-1mm}  \left[ \begin{array}{c} \textbf{b}_1 \\ \textbf{b}_2 \\ \vdots \\ \textbf{b}_V \end{array} \right] 
\label{eqn:stgcrf},
\end{equation}
where we group the variables by frames. 
Solving this system allows us to couple predictions $\mathbf{x}_v$ across all video frames $v \in \{1, \ldots, V\}$, positions, $p$ and labels $l$. If furthermore $A_v = A_v^T,\forall v$ and $T_{u,v} = T_{v,u}^T,\forall u,v$ then the resulting system is positive definite for any positive $\lambda$.

We now describe how the pairwise terms $A_{v},T_{u,v}$ are constructed through our CNN, and then discuss acceleration of the linear system in \refeq{eqn:stgcrf} by exploiting its structure.


\label{sec:gcrf}
\vspace{2mm}
\textbf{Spatial Connections:}
We define the spatial pairwise terms in terms of inner products of pixel-wise embeddings, as in \cite{gcrf2}.
At frame $v$ we couple the scores for a pair of patches $p_i, p_j$
taking the labels $l_m,l_n$ respectively as follows: 
\begin{eqnarray}
A_{v , {p_i,p_j}}\left(l_m,l_n\right) = \langle \mathcal{A}_{v, {p_i}}^{l_m},\mathcal{A}_{v, {p_j}}^{l_n}\rangle,
\label{eqn:pws}
\end{eqnarray}
where $i,j \in \{1,\ldots,P\}$ and $m,n \in \{ 1,\ldots,L \}$,  $v \in \{1,\ldots,V\}$, and $\mathcal{A}_{v, {p_j}}^{l_n}\in \mathbb{R}^D$ is the embedding associated to point $p_j$. 
In \refeq{eqn:pws} the $\mathcal{A}_{v, {p_j}}^{l_n}$ terms are image-dependent and delivered by a fully-convolutional ``embedding'' branch that feeds from the same CNN backbone architecture, and is denoted by $\mathcal{A}_v$ in \reffig{fig:temporalGCRF}. 

The implication of this form is that we can afford inference with a fully-connected graph. In particular the rank of the block matrix $A_v = \mathcal{A}_v^\top \mathcal{A}_v$, equals the embedding dimension $D$, which means that both the memory- and time- complexity of solving the linear system drops from $O(N^2)$ to $O(ND)$, which can be several orders of magnitude smaller. 
Thus, $\mathcal{A}_v \in \mathbb{R}^{N\times D}$







\vspace{2mm}
\textbf{Temporal Connections:} Turning to the \emph{temporal} pairwise terms, we couple patches $p_i,p_j$ coming from different frames $u,v$ taking the labels $l_m,l_n$ respectively as
\begin{eqnarray}
T_{u,v,{p_i,p_j}}\left(l_m,l_n\right) = \langle \mathcal{T}_{u,{p_i}}^{l_m},\mathcal{T}_{v, {p_j}}^{l_n}\rangle,
\label{eqn:pwt}
\end{eqnarray}
where $u,v \in \{1,\ldots,V\}$. The respective embedding terms are delivered by a  branch of the network that is separate, temporal embedding network denoted by  $\mathcal{T}_v$ in \reffig{fig:temporalGCRF}.

\begin{figure}[h!]
\centering
\includegraphics[width=\linewidth]{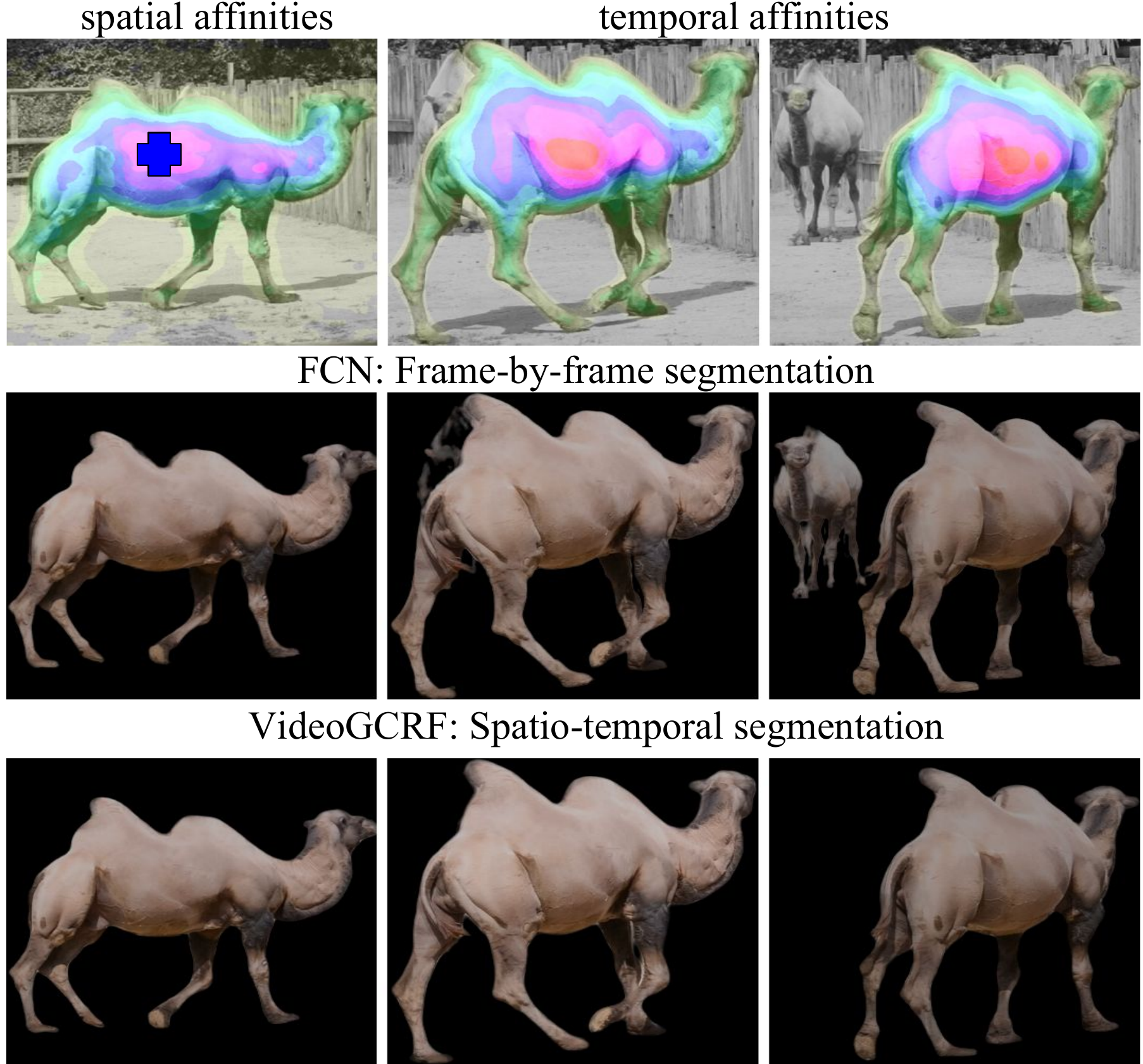}\\
\caption{Visualization of instance segmentation through VideoGCRF: In row 1 we focus on a single point of the CRF graph, shown as a cross, and show as a heatmap its spatial (inter-frame) and temporal (intra-frame) affinities to all other graph nodes. These correspond to a single column of the linear system in \refeq{eqn:stgcrf}. In row 2 we show the predictions that would be obtained by frame-by-frame segmentation,  relying exclusively on the FCN's unary terms, while in row 3 we show the results obtained after solving the VideoGCRF inference problem. We observe that in frame-by-frame segmentation a second camel is incorrectly detected due to its similar appearance properties. However, VideoGCRF inference  exploits temporal context and  focuses solely on the correct object.} \label{fig:emb} 
\end{figure}

In short, both the spatial pairwise and the temporal pairwise terms are composed as Gram matrices of spatial and temporal embeddings as $A_v = \mathcal{A}_{v}^\top \mathcal{A}_{v}$, and $T_{u,v} = \mathcal{T}_{u}^\top \mathcal{T}_{v}$. We visualize our spatio-temporal pairwise terms in \reffig{fig:emb}.

\vspace{2mm}
\textbf{VideoGCRF in Deep Learning:}
Our proposed spatio-temporal Gaussian CRF (VideoGCRF) can be viewed as generic deep learning modules for spatio-temporal
structured prediction, and as such can be plugged in at any stage of a deep
learning pipeline: either as the last layer, i.e. classifier, as in our semantic segmentation experiments (\refsec{sec:expcamvid}), or even in the low-level feature learning stage,
as in our instance segmentation experiments (\refsec{sec:expabl}).

\newcommand{\mycomment}[1]{}

\subsection{Efficient Conjugate-Gradient Implementation}
\label{sec:cg}
 We now describe an efficient implementation of the
conjugate gradient method  \cite{conjugategradient}, described in Algorithm~\ref{alg:cg}
that is customized for our VideoGCRFs. 


\begin{algorithm}
\caption{Conjugate Gradient Algorithm}\label{alg:cg}
\begin{algorithmic}[1]
\Procedure{ConjugateGradient}{}
\State \textbf{Input:} $\mathbf{A}$, $\mathbf{B}$, $\mathbf{x}_0$ \hspace{2mm} \textbf{Output:} $\mathbf{x} \mid \mathbf{A} \mathbf{x} = \mathbf{B}$
\State $\mathbf{r}_0 := \mathbf{B} - \mathbf{A x}_0$; \hspace{2mm} $\mathbf{p}_0 := \mathbf{r}_0$; \hspace{2mm} $k := 0$
\State $\text{repeat} $
\State $\qquad \alpha_k := \frac{\mathbf{r}_k^\mathsf{T} \mathbf{r}_k}{\mathbf{p}_k^\mathsf{T} \mathbf{A p}_k}  $
\State $\qquad \mathbf{x}_{k+1} := \mathbf{x}_k + \alpha_k \mathbf{p}_k $
\State $\qquad \mathbf{r}_{k+1} := \mathbf{r}_k - \alpha_k \mathbf{A p}_k $
\State $\qquad \hbox{if } \|\mathbf{r}_{k+1}\| \text{ is sufficiently small, then exit loop} $
\State $\qquad \beta_k := \frac{\mathbf{r}_{k+1}^\mathsf{T} \mathbf{r}_{k+1}}{\mathbf{r}_k^\mathsf{T} \mathbf{r}_k} $
\State $\qquad \mathbf{p}_{k+1} := \mathbf{r}_{k+1} + \beta_k \mathbf{p}_k $
\State $\qquad k := k + 1 $
\State $\text{end repeat} $
\State $\mathbf{x} = \mathbf{x}_{k+1}$
\EndProcedure
\end{algorithmic}
\end{algorithm}

The computational complexity of the conjugate gradient algorithm is determined by  the computation of the matrix-vector product $\mathbf{q} = A \mathbf{p}$, corresponding to \texttt{line :7} of Algorithm~\ref{alg:cg} (we drop the subscript $k$ for convenience).

We now discuss how to efficiently compute  $\mathbf{q}$ in a manner that is customized for this work.
In our case, the matrix-vector product $\mathbf{q} = A\mathbf{p}$ is expressed in terms of the spatial ($\mathcal{A}$) and temporal ($\mathcal{T}$) embeddings as follows: 

\begin{equation}
\resizebox{0.98\linewidth}{!}
{
$\left[ \begin{array}{c} \textbf{q}_1 \\ \textbf{q}_2 \\ \vdots \\ \textbf{q}_V \end{array} \right] \hspace{-2mm} = \hspace{-2mm}  
\begin{bmatrix} \mathcal{A}_{1}^T \mathcal{A}_{1} + \lambda \mathbf{I} & \mathcal{T}_{1}^T \mathcal{T}_{2} & \cdots & \mathcal{T}_{1}^T \mathcal{T}_{V}\\ \mathcal{T}_{2}^T \mathcal{T}_{1} & \mathcal{A}_{2}^T \mathcal{A}_{2} + \lambda \mathbf{I} & \cdots & \mathcal{T}_{2}^T \mathcal{T}_{V}  \\  &  & \vdots  &  \\ \mathcal{T}_{V}^T \mathcal{T}_{1} & \mathcal{T}_{V}^T \mathcal{T}_{2} & \cdots & \mathcal{A}_{V}^T \mathcal{A}_{V} + \lambda \mathbf{I} \end{bmatrix} \hspace{-2mm} \left[ \begin{array}{c} \textbf{p}_1 \\ \textbf{p}_2 \\ \vdots \\ \textbf{p}_V \end{array} \right]$
}
\label{eqn:stgcrf2}
\end{equation}

From \refeq{eqn:stgcrf2}, we can express $\mathbf{q}_i$ as follows:
\begin{equation}
\mathbf{q}_i =  \mathcal{A}_{i}^T \mathcal{A}_{i} \mathbf{p}_i + \lambda \mathbf{p}_i + \sum_{j \ne i} \mathcal{T}_i^T \mathcal{T}_j \mathbf{p}_j \text{.}
\label{eqn:qi}
\end{equation}

One optimization that we exploit in computing $\mathbf{q}_i$ efficiently is that we do not `explicitly' compute the matrix-matrix products $\mathcal{A}_{i}^T \mathcal{A}_{i}$ or $\mathcal{T}_i^T \mathcal{T}_j$. We note that $\mathcal{A}_{i}^T \mathcal{A}_{i} \mathbf{p}_i$ can be decomposed into two matrix-vector products as $\mathcal{A}_{i}^T \left( \mathcal{A}_{i} \mathbf{p}_i \right)$, where the expression in the brackets is evaluated first and yields a vector, which can then be multiplied with the matrix outside the brackets. This simplification alleviates the need to keep $N \times N$ terms in memory, and is computationally cheaper.
 
Further, from \refeq{eqn:qi}, we note that computation of $\mathbf{q}_i$ requires the matrix-vector product $\mathcal{T}_j \mathbf{p}_j \hspace{2mm} \forall j \ne i$. A \emph{black-box} implementation would therefore involve redundant computations, which we eliminate  by rewriting \refeq{eqn:qi} as:
\begin{equation}
\mathbf{q}_i =  \mathcal{A}_{i}^T \mathcal{A}_{i} \mathbf{p}_i + \lambda \mathbf{p}_i + \mathcal{T}_i^T \left( (\sum_{j}  \mathcal{T}_j \mathbf{p}_j) - \mathcal{T}_i \mathbf{p}_i \right) \text{.}
\label{eqn:qi2}
\end{equation}
This rephrasing allows us to precompute and cache $\sum_{j}  \mathcal{T}_j \mathbf{p}_j$, thereby eliminating redundant calculations. 

While so far we have assumed dense connections between the image frames, if we have sparse temporal connections (\refsec{sec:expabl}), i.e. each frame is connected to a subset of neighbouring frames in the temporal domain, the linear system matrix $A$ is sparse, and $\mathbf{q}_i$ is written as
\begin{equation}
\mathbf{q}_i =  \mathcal{A}_{i}^T \mathcal{A}_{i} \mathbf{p}_i + \lambda \mathbf{p}_i + \sum_{j \in \mathcal{N}(i)} \mathcal{T}_i^T \mathcal{T}_j \mathbf{p}_j \text{,}
\label{eqn:qi_sparse}
\end{equation}
where $\mathcal{N}(i)$ denotes the temporal neighbourhood of frame $i$. For very sparse connections caching may not be necessary because these involve little or no redundant computations.

\subsection{Backward Pass}
Since we rely on the Gaussian CRF we can get the back-propagation equation for the gradient of the loss with respect to the unary terms, $\mathbf{b}_v$, and the spatial/temporal embedding terms $\mathcal{A}_v,\mathcal{T}_v$ in closed form. Thanks to this we do not have to perform back-propagation in time which was needed e.g. in \cite{crfrnn} for DenseCRF inference. Following \cite{gcrf2}, 
the gradients of the unary terms $\frac{\partial \mathcal{L}}{\partial \mathbf{b}_v}$ are obtained from the solution of the following system:
\begin{equation} 
\begin{bmatrix} A_1 + \lambda \mathbf{I} & T_{1,2} & \cdots & T_{1,V}\\ T_{2,1} & A_2 + \lambda \mathbf{I} & \cdots & T_{2,V}  \\  &  & \vdots  &  \\ T_{V,1} & T_{V,2} & \cdots & A_V + \lambda \mathbf{I} \end{bmatrix} \hspace{-2mm} \left[ \begin{array}{c} \frac{\partial \mathcal{L}}{\partial \mathbf{b}_1} \\ \frac{\partial \mathcal{L}}{\partial \mathbf{b}_2} \\ \vdots \\ \frac{\partial \mathcal{L}}{\partial \mathbf{b}_V} \end{array} \right] \hspace{-1mm} = \hspace{-1mm}  \left[ \begin{array}{c} \frac{\partial \mathcal{L}}{\partial \mathbf{x}_1} \\ \frac{\partial \mathcal{L}}{\partial \mathbf{x}_2} \\ \vdots \\ \frac{\partial \mathcal{L}}{\partial \mathbf{x}_V} \end{array} \right] 
\label{eqn:grad_b}
\end{equation}

Once these are computed, the gradients of the spatial embeddings can be computed as follows:
\begin{equation}
 \frac{\partial \mathcal{L}}{\partial \mathcal{A}_v} = - \left(\frac{\partial \mathcal{L}}{\partial \mathbf{b}_v}  \otimes \mathbf{x}_v\right)\left( \left({\bf I} \otimes \mathcal{A}_v^\top\right) + \left(\mathcal{A}_v^\top \otimes {\bf I}\right) {Q_{D,N}}\right) \label{eqn2:grad_s}
 \end{equation}
while the gradients of the temporal embeddings are given by the following form:
\begin{equation}
 \frac{\partial \mathcal{L}}{\partial \mathcal{T}_v} = - \sum_{u} \left(\frac{\partial \mathcal{L}}{\partial \mathbf{b}_u}  \otimes \mathbf{x}_v\right)\left( \left({\bf I} \otimes \mathcal{T}_u^\top\right) + \left(\mathcal{T}_u^\top \otimes {\bf I}\right) {Q_{D,N}}\right) \label{eqn2:grad_s}
 \end{equation}
 where $Q_{D,N}$ is a permutation matrix, as in \cite{gcrf2}. 

\mycomment{
considering that the G-CRF \emph{layer} obtains a gradient for the loss $\mathcal{L}$ with respect to its output $\textbf{x}$,
$\frac{\partial \mathcal{L}}{\partial \textbf{x}}$,
the gradients of
the unary terms $ \frac{\partial \mathcal{L}}{\partial B}$ may be obtained by solving a new system of linear equations:

\begin{equation}
 ( A + \lambda \mathbf{I} ) \frac{\partial \mathcal{L}}{\partial B} = \frac{\partial \mathcal{L}}{\partial \textbf{x}}\text{,}
 \label{eqn2:dldbres}
\end{equation}
%
while  the gradients of the pairwise terms $ \frac{\partial \mathcal{L}}{\partial A}$ are given by:
\begin{equation}
 \frac{\partial \mathcal{L}}{\partial A} = - \frac{\partial \mathcal{L}}{\partial B}  \otimes \mathbf{x}\text{,}
 \label{eqn2:dlda2_1}
\end{equation}
where $\otimes$ denotes the Kronecker product operator.

As in \cite{gcrf}, the gradients of the unary terms $B$ can be computed by solving a system of linear equations,
\begin{equation}
 ( \mathcal{A}^\top\mathcal{A} + \lambda \mathbf{I} ) \frac{\partial \mathcal{L}}{\partial B} = \frac{\partial \mathcal{L}}{\partial \textbf{x}}\text{,}
 \label{eqn2:kerneldldbres}
\end{equation}

while the gradients of the pixel embeddings
 $\mathcal{A}$, are obtained from 
\begin{equation}
 \frac{\partial \mathcal{L}}{\partial \mathcal{A}} = - \left(\frac{\partial \mathcal{L}}{\partial B}  \otimes \mathbf{x}\right)\left( \left({\bf I} \otimes \mathcal{A}^\top\right) + \left(\mathcal{A}^\top \otimes {\bf I}\right) Q_{D,N}\right), \label{eqn2:kerneldlda}
 \end{equation}

where {$Q_{m,n}$} is a permutation matrix of size $mn \times mn$ defined, as in \cite{matcalc}, as follows:
\begin{equation}
 Q_{m,n} \text{vec}(M) = \text{vec}(M^\top)\text{,}
\end{equation}
where vec$(M)$ is the vectorization operator that vectorizes a matrix $M$ by stacking its columns.
When premultiplied with another matrix, $Q_{m,n}$  rearranges the ordering of rows of that matrix, while when postmultiplied with another matrix, $Q_{m,n}$ rearranges its columns.
}

\subsection{Implementation and Inference Time}
Our implementation is GPU based and exploits fast \emph{CUDA-BLAS}
linear algebra routines. It is implemented as a module in the Caffe2 library.
For spatial and temporal embeddings of size $128$, $12$ classes (\refsec{sec:expcamvid}), a $321 \times 321$ input image, and network stride of $8$, our $2,3,4$ frame inferences take $0.032$s, $0.045$s and $0.061$s on average respectively. Without the caching procedure described in \refsec{sec:cg}, the $4$ frame inference takes $0.080$s on average. This is orders of magnitude faster than the DenseCRF method \cite{densecrf} which takes $0.2$s on average for spatial CRF for a single input frame. These timing statistics were estimated on a \texttt{GTX-1080} GPU.

\newcommand{\myhgt}{.55\linewidth}

\section{Experiments}
\label{sec:exp}
\begin{figure*}
    \centering
    \begin{minipage}[c]{0.29\textwidth}
    \centering
    \includegraphics[width=0.9\linewidth]{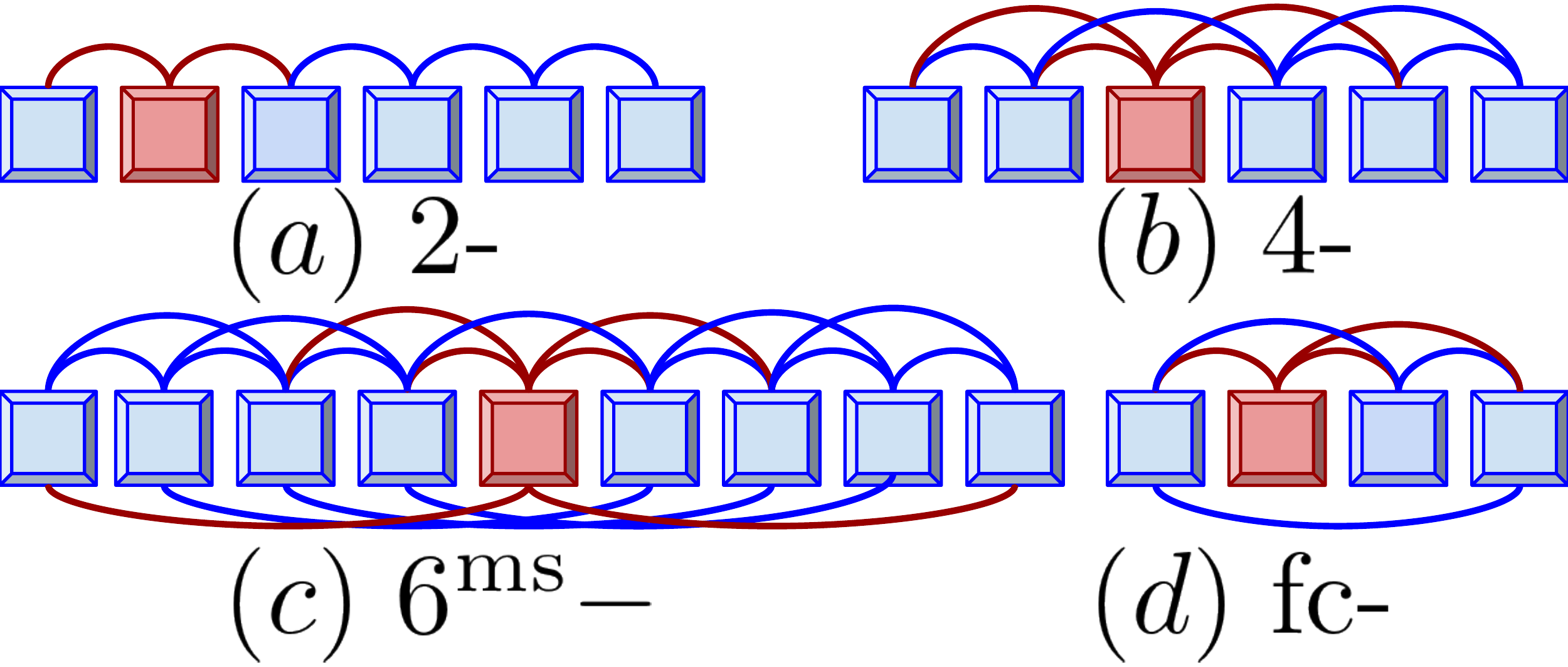}
    \caption{Temporal neighbourhoods 
    in our ablation study: boxes denote video frames and the arcs
    connecting them are pairwise connections. The frame in red has all neighbours present in the temporal context.
    }
    \label{fig:tempnbh}
    \end{minipage}\hfill
    \begin{minipage}[c]{0.7\textwidth}
    \centering
    \includegraphics[width=0.95\linewidth]{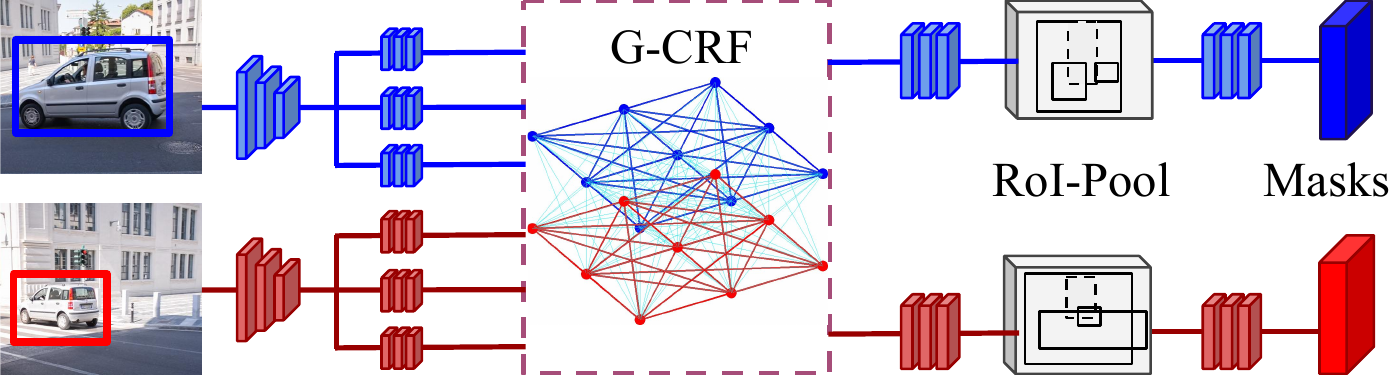}
    \caption{Spatio-temporal structured prediction in Mask-RCNN. Here we use CRFs in the feature learning stage before the ROI-Pooling (and not as the final classifier). This helps learn
    mid-level features which are better aware of the spatio-temporal context.}
    \label{fig:maskrcnn} 
    \end{minipage}
\end{figure*}

\noindent \textbf{Experimental Setup.}
We describe the basic setup followed for our experiments.
As in \cite{gcrf2}, we use a $3-$phase training strategy for our methods. We first train the unary network without the spatio-temporal embeddings. We next train the subnetwork delivering the spatio-temporal embeddings
with the softmax cross-entropy loss to enforce the following objectives: $A_{p_1,p_2}\left(l_1,l_2\right) < A_{p_1,p_2}\left(l'_1\neq l_1,l'_2\neq l_2\right)$, and 
$T_{u,v,p_1,p_2}\left(l_1,l_2\right) < T_{u,v,p_1,p_2}\left(l'_1\neq l_1,l'_2\neq l_2\right)$, where $l_1,l_2$
are the ground truth labels for pixels $p_1,p_2$.
Finally, we combine the unary and pairwise networks, and train them together in end-to-end fashion. 
Unless otherwise stated, we use stochastic gradient descent to train our networks with a momentum of $0.9$ and a weight decay of $5e^{-4}$. For segmentation experiments, we use a base-learning rate of $2.5e^{-3}$ for training the unaries, $2.5e^{-4}$ for training the embeddings, and $1e^{-4}$ for finetuning the unary and embeddings together, using a polynomial-decay with power of $0.9$. For the instance segmentation network, we use a single stage training for the unary and pairwise streams: we train the network for $16$K iterations, with a base learning rate  
 of $0.01$ which is reduced to $0.001$ after $12$K iterations. The weight decay is $1e^{-4}$. For our instance tracking experiments, we use unaries from \cite{onavos} and do not refine them, rather use them as an input to our network.
 We employ horizontal flipping and scaling by factors between $0.5$ and $1.5$ during training/testing for all methods, except in the case of instance segmentation experiments (\refsec{sec:expabl}).

\vspace{2mm}
\noindent \textbf{Datasets.}
We use the three datasets for our experiments:

\emph{DAVIS.} The DAVIS dataset \cite{davis2} consists of $30$ training and $20$ validation videos containing $2079$ and $1376$ frames respectively. Each video comes with manually annotated segmentation masks for foreground object instances.

\emph{DAVIS-Person.} While the DAVIS dataset \cite{davis1} provides densely annotated frames for instance segmentation, it lacks object category labels. For category prediction
tasks such as semantic and instance segmentation, we create a subset of the {DAVIS} dataset containing videos from the category person. By means of visual inspection, we select $35$ and $18$ video sequences from the training and validation sets respectively containing $2463$ training and $1182$ validation images, each containing at least one person. Since the DAVIS dataset comes with only the \emph{foreground} instances labeled, we manually annotate the image regions containing \emph{unannotated person} instances with the \emph{do-not-care} label. These image regions do not participate in the training or the evaluation. We call this the DAVIS-person dataset.

\emph{CamVid.} The CamVid dataset \cite{camvid,camvid2}, is a dataset containing videos of driving scenarios for urban scene understanding. It comes with $701$ images annotated with pixel-level category labels at $1$ fps. Although the original
dataset comes with $32$ class-labels, as in \cite{SegNet2015,fso,tiramisu}, we predict $11$ semantic classes and use the train-val-test split of $367$, $101$ and $233$ frames respectively.


\subsection{Ablation Study on Semantic and Instance Segmentation Tasks}
\label{sec:expabl}
In these experiments, we use the DAVIS Person dataset described in \refsec{sec:exp}. The aim here is to explore the various
design choices available to us when designing networks for spatio-temporal structured prediction for semantic segmentation, and proposal-based instance segmentation tasks.

\vspace{2mm}
\noindent \textbf{Semantic Segmentation Experiments.} 
Our first set of experiments studies the effect of varying the sizes of the spatial and temporal embeddings, the degree of the temporal connections, and multi-scale temporal connections for VideoGCRF. For these set of experiments, our baseline network, or \emph{base-net} is a single resolution ResNet-101 network, with altered network strides as in \cite{deeplabv2} to produce a spatial down-sampling factor of $8$. 
The evaluation metric used is the mean pixel Intersection over Union (IoU).

In \reftab{table:sizeembed} we study the effect of varying the sizes of the spatial and temporal embeddings for $2-$frame inference. Our best results are achieved at spatio-temporal embeddings of size $128$. The improvement over the base-net is $4.2\%$. In subsequent experiments we fix the size of our embeddings to $128$. We next study the effect of varying the size of the temporal context and temporal neighbourhoods. The temporal context is defined as the number of video frames $\mathcal{V}$ which are considered simultaneously in one linear system (\refeq{eqn:stgcrf}). The temporal context $\mathcal{V}$ is limited by the GPU RAM: for a ResNet-101 network, an input image of size $321 \times 321$, embeddings of size $128$, we can currently fit $\mathcal{V} = 7$ frames on $12$ GB of GPU RAM. Since $\mathcal{V}$ is smaller than the number of frames in the video, we divide the video into overlapping sets of $\mathcal{V}$ frames, and average the predictions for the common frames.

The temporal neighbourhood for a frame (\reffig{fig:tempnbh}) is defined as the number of frames it is directly connected to via pairwise connections. A fully connected neighbourhood ({fc$-$}) is one in which there are pairwise terms between every pair of frames available in the temporal context. We experiment with $2-$, $4-$, multiscale $6^{\text{ms}}-$ and {fc$-$} connections. The $6^{\text{ms}}-$ neighbourhood connects a frame to neighbours at distances of $2^0$, $2^1$ and $2^2$ (or $1,2,4$) frames on either side. \reftab{table:sizetemporal} reports our results for different combinations of temporal neighbourhood and context. It can be seen that dense connections improve performance for smaller temporal contexts, but for a temporal context of $7$ frames, an increase in the complexity of temporal connections leads to a moderate decrease in performance. This could be a consequence of
the long-range interactions having the same weight as short-range
interactions. In the future we intend to mitigate this issue by complementing our embeddings with the temporal distance between frames.

\begin{table}[!h]
\centering
\scalebox{0.95}{
\begin{tabular}{l |c|c|c|c}
\hline
base-net  &\multicolumn{4}{c}{$\textbf{81.16}$} \\ \hline
\hline
{VideoGCRF}&\multicolumn{4}{c}{{spatial dimension$\rightarrow$}}\\ \cline{1-1}
temporal dimension$\downarrow$& {64} & {128} & {256} & {512} \\ \hline
{64} & $84.89$& $85.21$ & $85.20$ & $84.98$\\ \hline
{128} & $85.18$ & $\mathbf{86.38}$ & $86.34$ & $84.91$\\ \hline
{256} & $85.92$ & $86.37$ & $85.95$ & $84.92$\\ \hline
{512} & $84.85$& $85.95$ & $84.95$ & $84.21$\\ \hline
\end{tabular}}
\caption{Ablation study: mean IoU on the DAVIS-person dataset using $2$ frame fc$-$ connections. We study the effect of varying the size of the spatial \& temporal embeddings.\label{table:sizeembed}}
\end{table}

\begin{table}[!h]
\centering
\begin{tabular}{l |c|c|c|c}
\hline
base-net  &\multicolumn{4}{c}{$\textbf{81.16}$} \\ \hline
\hline
{VideoGCRF}&\multicolumn{4}{c}{{temporal neighbourhood $\rightarrow$}}\\ \cline{1-1}
temporal context$\downarrow$& {$2-$} & {$4-$} & {$6^{\text{ms}}-$} & {fc$-$} \\ \hline
{2} & $-$& $-$ & $-$ & $86.38$\\ \hline
{3} & $86.42$ & $-$ & $-$ & $86.51$\\ \hline
{4} & $86.70$ & $-$ & $-$ & $86.82$\\ \hline
{7} & $\mathbf{86.98}$& $86.79$ & $86.82$ & $86.42$\\ \hline
\end{tabular}
\caption{Ablation study: mean IoU on the DAVIS-person dataset. Here we study the effect of varying the size of the temporal context and neighbourhood.\label{table:sizetemporal}}
\end{table}

\vspace{2mm}
\noindent \textbf{Instance Segmentation Experiments.}
We now demonstrate the utility of our VideoGCRF method for the task of proposal-based instance segmentation.
Our hypothesis is that coupling  predictions across frames is advantageous for instance segmentation methods. We actually show that the performance
of the instance segmentation methods improves as we increase the temporal context via VideoGCRF, and obtain our best results with fully-connected temporal neighbourhoods.
Our baseline for this task is the Mask-RCNN framework of \cite{he2017maskrcnn} using the ResNet-50 network as the convolutional \emph{body}. The Mask-RCNN framework uses precomputed
bounding box proposals for this task. It computes convolutional features on the input image using the convolutional \emph{body} network, crops out the features corresponding to 
image regions in the proposed bounding boxes via Region-Of-Interest (RoI) pooling, and then has $3$ \emph{head} networks to predict (i) class scores and bounding box regression parameters, (ii) keypoint locations, and (iii) instance masks. Structured prediction coupling the predictions of all the proposals over all the video frames is a computationally challenging task, since typically we have $100-1000$s of proposals per image, and it is not obvious which proposals from one frame should influence which proposals in the other frame. To circumvent this issue, we use our VideoGCRF before the RoI pooling stage as shown in \reffig{fig:maskrcnn}. Instead of coupling final predictions, we thereby couple mid-level features over the video frames, thereby improving the features which are ultimately  used to make predictions.

\begin{figure*}[h!]
\centering
\includegraphics[width=0.75\linewidth]{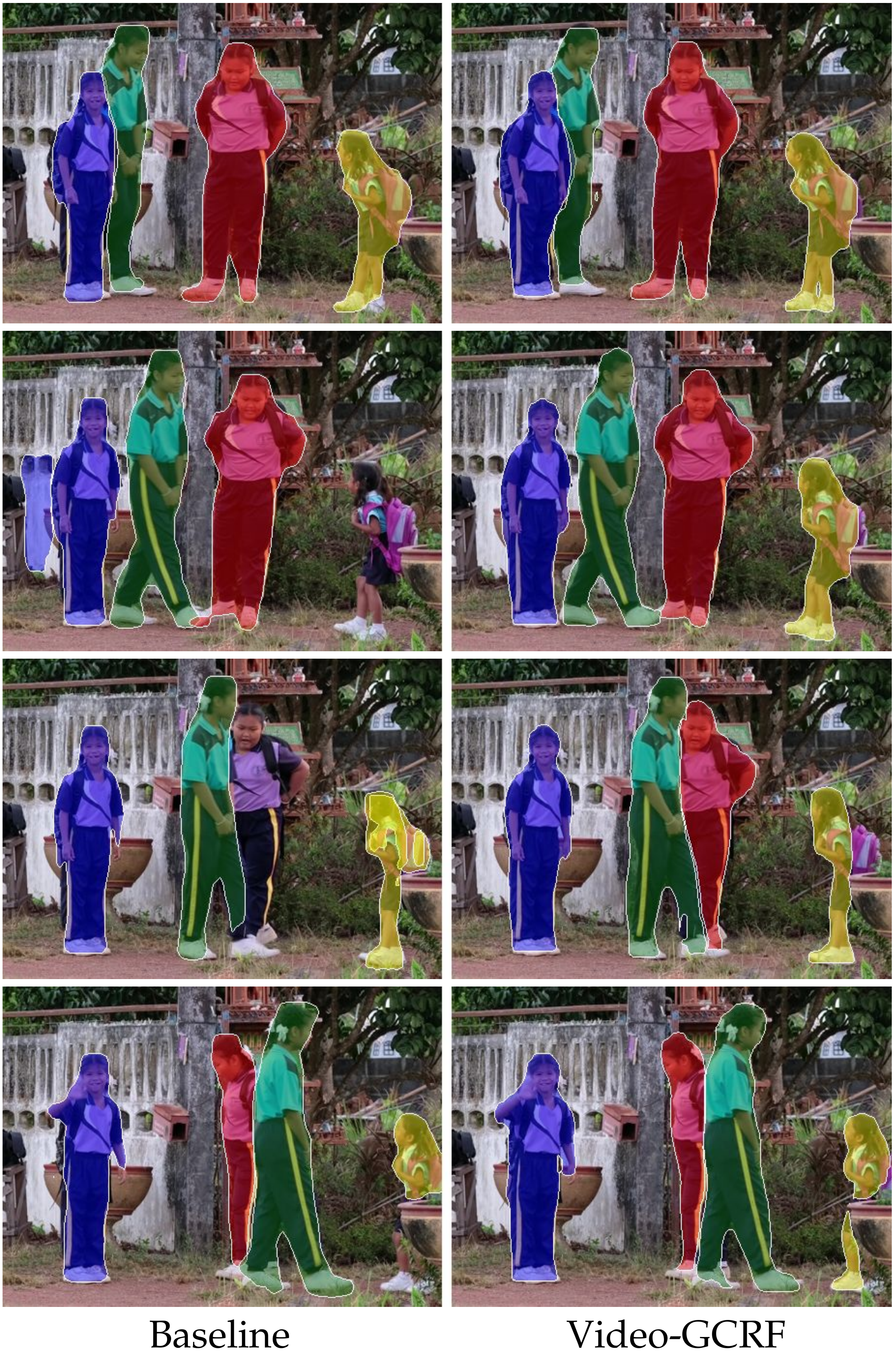}
\caption{Instance Segmentation results on the DAVIS Person Dataset. We observe that prediction based on unary terms alone leads to missing instances and some false predictions. These errors are corrected by VideoGCRFs, which smooth the predictions by taking into account the temporal context.}
\label{fig:instances} 
\end{figure*}

For evaluation, we use the standard COCO performance metrics: $\text{AP}_{50}$, $\text{AP}_{75}$, and AP (averaged
over IoU thresholds), evaluated using mask IoU. \reftab{table:maskrcnn} reports our instance segmentation results. We note that the performance of the Mask-RCNN framework
increases consistently as we increase the temporal context for predictions. Qualitative results are available in \reffig{fig:instances}.

\begin{table}[h!]
\centering
\begin{tabular}{l|r|r|r}
\hline
{Method} & $\text{AP}_{50}$ & $\text{AP}_{75}$ & AP \\ \hline
{ResNet50-baseline} & 0.610 & 0.305 & 0.321 \\ 
{spatial CRF} \cite{gcrf2} &  0.618 & 0.310 & 0.329 \\
{2-frame VideoGCRF} &  0.619 & 0.310 & 0.331 \\
{3-frame VideoGCRF} &  0.631 & 0.321 & 0.330 \\
{4-frame VideoGCRF} &  0.647 & 0.336 & 0.349 \\
\hline
\end{tabular}
\caption{Instance Segmentation using ResNet-50 Mask R-CNN on the Davis Person Dataset}
\label{table:maskrcnn}
\end{table}

\begin{table}[h!]
\centering
\begin{tabular}{l|r}
\hline
{Method} & mean IoU\\ \hline
Mask Track \cite{masktrack} & 79.7 \\
OSVOS \cite{osvos} & 79.8 \\
{Online Adaptation \cite{onavos} } & 85.6 \\
{Online Adaptation + Spatial CRF \cite{gcrf2}} & {85.9} \\
{Online Adaptation + 2-Frame VideoGCRF} & {86.3} \\
{Online Adaptation + 3-Frame VideoGCRF} & {86.5} \\
\hline
\end{tabular}
\caption{Instance Tracking on the Davis val Dataset}
\label{tab:inst}
\end{table}

\subsection{Instance Tracking}
\label{sec:expdavis}
We use the DAVIS dataset described in \refsec{sec:exp}.
Instance tracking involves predicting foreground segmentation masks for each video frame given the foreground segmentation for the first video frame. 
We demonstrate that incorporating temporal context helps improve performance in instance tracking methods. To this end we extend the online adaptation approach of \cite{onavos} which is the state-of-the-art approach on the DAVIS benchmark with our VideoGCRF. We use their publicly available software based on the TensorFlow library to generate the unary terms for each of the frames in the video,
and keep them fixed. We use a ResNet-50 network to generate spatio-temporal embeddings and use these alongside the unaries computed from \cite{onavos}.
The results are reported in table \reftab{tab:inst}. We compare performance of VideoGCRF against that of just the unaries from \cite{onavos}, and also with spatial CRFs from \cite{gcrf2}. The evaluation criterion is the mean pixel-IoU. It can be seen that temporal context improves performance. We hypothesize that re-implementing the software from \cite{onavos} in Caffe2 and back-propagating on the unary branch of the network would yield further improvements.

\begin{table*}[h!]
\footnotesize
\centering
\setlength\tabcolsep{5.3pt} 
 \begin{tabular}{c | c | c | c | c | c | c | c | c | c | c | c || c }
Model &\rotatebox{90}{\scriptsize{Building}} & \rotatebox{90}{Tree} & \rotatebox{90}{Sky} & \rotatebox{90}{Car} & \rotatebox{90}{Sign} & \rotatebox{90}{Road} & \rotatebox{90}{\scriptsize{Pedestrian}} & \rotatebox{90}{Fence} & \rotatebox{90}{Pole} & \rotatebox{90}{\scriptsize{Sidewalk}} & \rotatebox{90}{Cyclist} & \rotatebox{90}{m-IoU} \\
\hline \hline
DeconvNet \cite{noh2015learning} & \multicolumn{11}{|c||}{$-$} &  $48.9$ \\
\hline
SegNet  \cite{SegNet2015}  & $68.7$ & $52.0$ & $87.0$ & $58.5$ & $13.4$ & $86.2$ & $25.3$ & $17.9$ & $16.0$ & $60.5$ & $24.8$ & $46.4$ \\
\hline
Bayesian SegNet \cite{KendallBC15} & \multicolumn{11}{|c||}{$-$} &  $63.1$ \\
\hline
Visin et al. \cite{VisinKCBMC15}   & \multicolumn{11}{|c||}{$-$} & $58.8$ \\
\hline
FCN8 \cite{fcnn} &   $77.8$ & $71.0$ & $88.7$ & $76.1$ & $32.7$ & $91.2$ & $41.7$ & $24.4$ & $19.9$ & $72.7$ & $31.0$ & $57.0$ \\
\hline
DeepLab-LFOV \cite{deeplab1}   & $81.5$ & $74.6$ & $89.0$ & $82.2$ & $42.3$ & $92.2$ & $48.4$ & $27.2$ & $14.3$ & $75.4$ & $50.1$ & $61.6$ \\
\hline
Dilation8  \cite{dilationkoltun}    & $82.6$ & $76.2$ & $89.0$ & $84.0$ & $46.9$ & $92.2$ & $56.3$ & $35.8$ & $23.4$ & $75.3$ & $55.5$ & $65.3$ \\
\hline
Dilation8 + FSO \cite{fso}  & ${84.0}$ & $77.2$ & $91.3$ & ${85.6}$ & ${49.9}$ & $92.5$ & $59.1$ & ${37.6}$ & $16.9$ & $76.0$ & ${57.2}$ & $66.1$ \\
\hline
Tiramisu \cite{tiramisu}& $83.0$ & ${77.3}$ & ${93.0}$ & $77.3$ & $43.9$ & $94.5$ & ${59.6}$ & $37.1$ & $37.8$ & ${82.2}$ & $50.5$ & ${66.9}$ \\
\hline
Gadde et al. \cite{videocnn}   & \multicolumn{11}{|c||}{$-$} & $67.1$ \\
\hline
\multicolumn{12}{c}{Results with our ResNet-101 Implementation} \\
\hline
Basenet ResNet-101 (Ours) & $81.2$ &  $75.1$ &  $90.3$ &  $85.2$ &   $48.3$ &  $93.9$ &  $57.7$ &  $39.9$ & $15.9$ &   $80.5$ &    $54.8$ & $65.7$ \\
Basenet + Spatial CRF \cite{gcrf2}  & $81.6$ & $75.7$ & $90.4$ & $86.8$ & $48.1$ & $94.0$ & $59.1$ & $39.2$ & $15.7$ & $80.7$ & $54.7$ & $66.0$\\
\hline
Basenet + 2-Frame VideoGCRF & $82.0$ & $76.1$ & $91.1$ & $86.2$ & $51.7$ & $93.8$ & $64.2$ & $24.5$ & $25.0$ & $80.1$ & $61.7$ & $66.9$ \\
Basenet + 3-Frame VideoGCRF & $82.1$ & $76.0$ & $91.1$ & $86.1$ & $52.0$ & $93.7$ & $64.5$ & $24.9$ & $24.4$ & $79.9$ & $61.8$ & $67.0$ \\
\hline
\multicolumn{12}{c}{Results after Cityscapes Pretraining} \\
\hline
Basenet ResNet-101 (Ours) &  $85.5$ & $77.4$ & $90.9$ & $88.4$ & $62.3$ & $95.4$ & $64.8$ & $62.1$ & $33.3$ & $85.5$ & $60.5$ & $73.3$ \\
Basenet + denseCRF post-processing \cite{densecrf}  & $84.3$ & $76.1$ & $90.5$ & $88.9$ & $65.1$ & $95.4$ & $65.4$ & $61.5$ & $34.1$ & $85.8$ & $66.2$ & $73.9$\\
Basenet + Spatial CRF \cite{gcrf2}               & $86.0$ & $77.8$ & $91.2$ & $90.8$ & $63.6$ & $95.9$ & $66.5$ & $61.2$ & $35.3$ & $86.9$ & $65.8$ & $74.6$\\
\hline
Basenet + 2-Frame VideoGCRF & $86.0$ & $78.3$ & $91.2$ & $92.0$ & $63.4$ & $96.3$ & $67.0$ & $62.5$ & $34.4$ & $87.7$ & $66.1$ & $75.0$ \\
Basenet + 3-Frame VideoGCRF & $86.1$ & $78.3$ & $91.2$ & $92.2$ & $63.7$ & $96.4$ & $67.3$ & $63.0$ & $34.4$ & $87.8$ & $66.4$ & $75.2$ \\

\hline
 \end{tabular}

 \caption{Results on CamVid dataset. We compare our results with some of the previously published methods, as well as our own implementation of the ResNet-101 network which serves as our base network.}
 \label{tab:CamVid}
 \setlength\tabcolsep{6pt} 
\end{table*}

\begin{figure*}[h!]
\includegraphics[width=\linewidth]{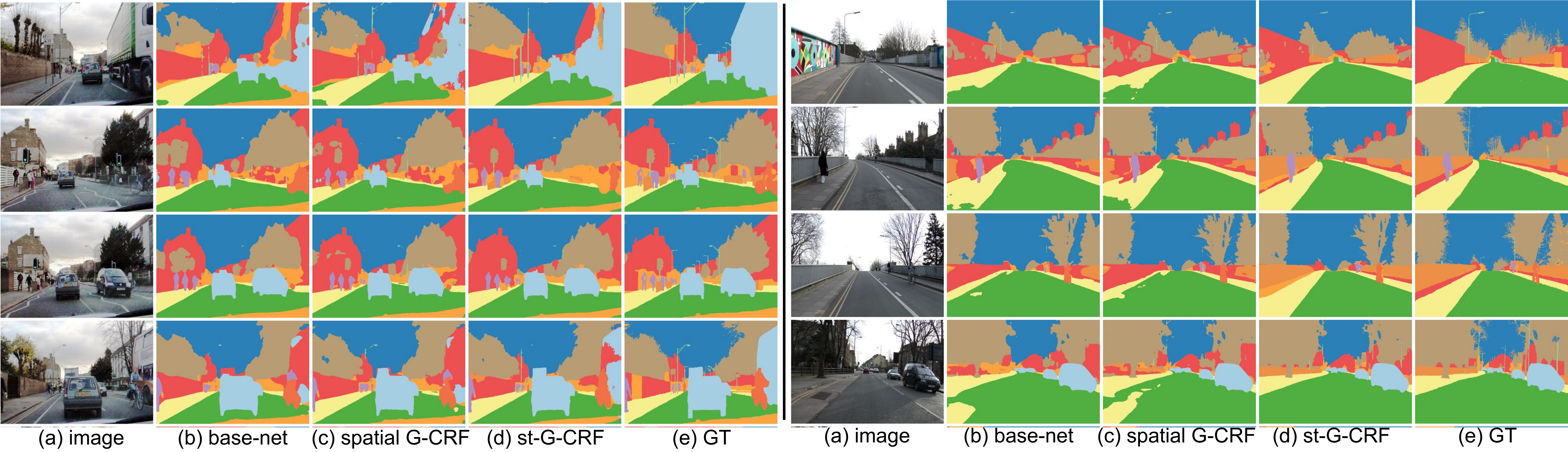}
\caption{Qualitative results on the CamVid dataset. We note that the temporal context from neighbouring frames helps improve the prediction of the truck on the right in the first video, and helps distinguish between the road and the pavement in the second video, overall giving us smoother predictions in both cases. }
\label{fig:camvid} 
\end{figure*}

\subsection{Semantic Segmentation on CamVid Dataset}
\label{sec:expcamvid}
We now employ our VideoGCRF for the task of semantic
video segmentation on the CamVid dataset. Our base network here is our own implementation of ResNet-101 with pyramid spatial pooling as in \cite{pspnet}. Additionally, we pretrain our networks on the Cityscapes dataset \cite{Cordts2016cityscapes}, and report results both with and without pretraining on Cityscapes. We report improvements over the baseline networks in both settings. Without pretraining, we see an improvement of $1.3\%$ over the base-net, and with pretraining we see an improvement of $1.9\%$. The qualitative results are shown in \reffig{fig:camvid}. We notice that VideoGCRF benefits from temporal context, yielding smoother predictions across video frames.

\section{Conclusion}
In this work, we propose VideoGCRF, an end-to-end trainable Gaussian CRF for efficient spatio-temporal structured prediction.
We empirically show performance improvements
on several benchmarks thanks to an increase of the temporal context.
This additional functionality comes at negligible computational overhead owing to efficient implementation and the strategies to eliminate redundant computations. In  future work we want to incorporate optical flow techniques in our framework as they provide a natural means to capture temporal
correspondence. Further, we also intend to use temporal distance between frames as an additional term in the expression of the pairwise interactions alongside dot-products of our embeddings. We would also like to use VideoGCRF for dense regression tasks such as depth estimation.
Finally, we believe that our method for spatio-temporal structured prediction can prove useful in the unsupervised and semi-supervised setting.

\appendix
\renewcommand{\thesubsection}{\Alph{subsection}}
\part*{Appendix}
\section*{Gradient Expressions for Spatio-Temporal G-CRF Parameters}
As described in the manuscript, to capture the spatio-temporal context, we propose two kinds of pairwise interactions: (a) pairwise terms between patches in the same frame (spatial pairwise terms), and (b) pairwise terms between patches in different frames (temporal pairwise terms).

Denoting  the spatial pairwise terms at frame $v$ by $A_{v}$ and the temporal pairwise terms between frames $u,v$ as  $T_{u,v}$, our inference equation is written as 
\begin{equation}
\begin{bmatrix} A_1 + \lambda \mathbf{I} & T_{1,2} & \cdots & T_{1,V}\\ T_{2,1} & A_2 + \lambda \mathbf{I} & \cdots & T_{2,V}  \\  &  & \vdots  &  \\ T_{V,1} & T_{V,2} & \cdots & A_V + \lambda \mathbf{I} \end{bmatrix} \hspace{-2mm} \left[ \begin{array}{c} \textbf{x}_1 \\ \textbf{x}_2 \\ \vdots \\ \textbf{x}_V \end{array} \right] \hspace{-1mm} = \hspace{-1mm}  \left[ \begin{array}{c} \textbf{b}_1 \\ \textbf{b}_2 \\ \vdots \\ \textbf{b}_V \end{array} \right] 
\label{eqn:sup_stgcrf},
\end{equation}
where we group the variables by frames. 
Solving this system allows us to couple predictions $\mathbf{x}_v$ across all video frames $v \in \{1, \ldots, V\}$, positions, $p$ and labels $l$. If furthermore $A_v = A_v',\forall v$ and $T_{u,v} = T_{v,u}',\forall u,v$ then the resulting system is positive definite for any positive $\lambda$.

As in the manuscript, at frame $v$ we couple the scores for a pair of patches $p_i, p_j$
taking the labels $l_m,l_n$ respectively as follows: 
\begin{eqnarray}
A_{v , {p_i,p_j}}\left(l_m,l_n\right) = \langle \mathcal{A}_{v, {p_i}}^{l_m},\mathcal{A}_{v, {p_j}}^{l_n}\rangle,
\label{eqn:supp_pws}
\end{eqnarray}
where $i,j \in \{1,\ldots,P\}$ and $m,n \in \{ 1,\ldots,L \}$,  $v \in \{1,\ldots,V\}$, and $\mathcal{A}_{v, {p_j}}^{l_n}\in R^D$ is the embedding associated to point $p_j$. 

Thus, $\mathcal{A}_v \in \mathbb{R}^{N\times D}$, where $N = P \times L$. Further, to design the \emph{temporal} pairwise terms, we couple patches $p_i,p_j$ coming from different frames $u,v$ taking the labels $l_m,l_n$ respectively as
\begin{eqnarray}
T_{u,v,{p_i,p_j}}\left(l_m,l_n\right) = \langle \mathcal{T}_{u,{p_i}}^{l_m},\mathcal{T}_{v, {p_j}}^{l_n}\rangle,
\label{eqn:supp_pwt}
\end{eqnarray}
where $u,v \in \{1,\ldots,V\}$.

In short, both the spatial pairwise and the temporal pairwise terms are composed as Gram matrices of spatial and temporal embeddings as $A_v = \mathcal{A}_{v}^\top \mathcal{A}_{v}$, and $T_{u,v} = \mathcal{T}_{u}^\top \mathcal{T}_{v}$.

Using the definitions from \refeq{eqn:supp_pws} and \refeq{eqn:supp_pwt}, we can rewrite the inference equation as
\begin{equation}
\resizebox{0.98\linewidth}{!}
{
$\begin{bmatrix} \mathcal{A}_{1}^T \mathcal{A}_{1} + \lambda \mathbf{I} & \mathcal{T}_{1}^T \mathcal{T}_{2} & \cdots & \mathcal{T}_{1}^T \mathcal{T}_{V}\\ \mathcal{T}_{2}^T \mathcal{T}_{1} & \mathcal{A}_{2}^T \mathcal{A}_{2} + \lambda \mathbf{I} & \cdots & \mathcal{T}_{2}^T \mathcal{T}_{V}  \\  &  & \vdots  &  \\ \mathcal{T}_{V}^T \mathcal{T}_{1} & \mathcal{T}_{V}^T \mathcal{T}_{2} & \cdots & \mathcal{A}_{V}^T \mathcal{A}_{V} + \lambda \mathbf{I} \end{bmatrix} \hspace{-2mm} \left[ \begin{array}{c} \textbf{x}_1 \\ \textbf{x}_2 \\ \vdots \\ \textbf{x}_V \end{array} \right]$
\hspace{-2mm} = \hspace{-2mm}  
$\left[ \begin{array}{c} \textbf{b}_1 \\ \textbf{b}_2 \\ \vdots \\ \textbf{b}_V \end{array} \right]$ 
}
\label{eqn:supp_stgcrf2}
\end{equation}

From \refeq{eqn:supp_stgcrf2}, we can express $\mathbf{b}_v$ as follows:
\begin{equation}
\mathbf{b}_v =  \mathcal{A}_{v}^T \mathcal{A}_{v} \mathbf{x}_v + \lambda \mathbf{x}_v + \sum_{u \ne v} \mathcal{T}_v^T \mathcal{T}_u \mathbf{x}_u \text{,}
\label{eqn:supp_qi}
\end{equation}

which can be compactly written as 
\begin{equation}
\mathbf{b}_v =  A_{v} \mathbf{x}_v + \lambda \mathbf{x}_v + \sum_{u \ne v} {T}_{v,u} \mathbf{x}_u \text{.}
\label{eqn:supp_qi2}
\end{equation}

We will use \refeq{eqn:supp_qi2} to derive gradient expressions for $\frac{\partial \mathcal{A}_v}{\partial \mathcal{L}}$ and $\frac{\partial \mathcal{T}_v}{\partial \mathcal{L}}$.

\subsection{Gradients of the Unary Terms}
As in \cite{gcrf,gcrf2}, the gradients of the unary terms $\frac{\partial \textbf{b}_v}{\partial \mathcal{L}}$ are obtained from the solution of the following
system of linear equations:
\begin{equation}
\resizebox{0.98\linewidth}{!}
{
$\begin{bmatrix} \mathcal{A}_{1}^T \mathcal{A}_{1} + \lambda \mathbf{I} & \mathcal{T}_{1}^T \mathcal{T}_{2} & \cdots & \mathcal{T}_{1}^T \mathcal{T}_{V}\\ \mathcal{T}_{2}^T \mathcal{T}_{1} & \mathcal{A}_{2}^T \mathcal{A}_{2} + \lambda \mathbf{I} & \cdots & \mathcal{T}_{2}^T \mathcal{T}_{V}  \\  &  & \vdots  &  \\ \mathcal{T}_{V}^T \mathcal{T}_{1} & \mathcal{T}_{V}^T \mathcal{T}_{2} & \cdots & \mathcal{A}_{V}^T \mathcal{A}_{V} + \lambda \mathbf{I} \end{bmatrix} \hspace{-2mm} \left[ \begin{array}{c} \frac{\partial \mathcal{L}}{\partial \textbf{b}_1} \\ \frac{\partial \mathcal{L}}{\partial \textbf{b}_2} \\ \vdots \\ \frac{\partial \mathcal{L}}{\partial \textbf{b}_V} \end{array} \right]$
\hspace{-2mm} = \hspace{-2mm}  
$\left[ \begin{array}{c} \frac{\partial \mathcal{L}}{\partial \textbf{x}_1} \\ \frac{\partial \mathcal{L}}{\partial \textbf{x}_2} \\ \vdots \\ \frac{\partial \mathcal{L}}{\partial \textbf{x}_V} \end{array} \right]$,
}
\label{eqn:supp_unarygrad}
\end{equation}

where $\mathcal{L}$ is the network loss. Once we have $\frac{\partial \mathcal{L}}{\partial \textbf{b}_v}$, we use it to compute the gradients of the spatio-temporal embeddings.

\subsection{Gradients of the Spatial Embeddings}
We begin with the observation that computing $\frac{\partial \mathcal{A}_v}{\partial \mathcal{L}}$ requires us to first derive the expression for $\frac{\partial {A}_v}{\partial \mathcal{L}}$. To this end, we ignore terms from \refeq{eqn:supp_qi2} that do not depend on $\mathbf{b}_v$ or $A_v$ and write it as $\mathbf{b}_v =  A_{v} \mathbf{x}_v + c$.
We now use the result from \cite{gcrf,gcrf2} that when 
\begin{equation*}
A_v \mathbf{x}_v = \mathbf{b}_v,
\end{equation*}
the gradients of $A_v$ are expressed as
\begin{equation}
\frac{\partial \mathcal{L}}{\partial A_v} = - \frac{\partial \mathcal{L}}{\partial \mathbf{b}_v} \otimes \mathbf{x}_v,
\label{eqn:supp_vid}
\end{equation}
where $\otimes$ denotes the Kronecker product operator.

To compute $\frac{\partial \mathcal{A}_v}{\partial \mathcal{L}}$, we use the chain rule of differentiation as follows:
\begin{equation}
 \frac{\partial \mathcal{L}}{\partial \mathcal{A}_v} =  \left(\frac{\partial \mathcal{L}}{\partial A_v}\right)  \left(\frac{\partial A_v}{\partial \mathcal{A}_v}\right) = \left(\frac{\partial \mathcal{L}}{\partial A_v}\right)  \left(\frac{\partial }{\partial \mathcal{A}_v}\mathcal{A}_v^T \mathcal{A}_v
 \right)\text{,}\label{eqn2:kernelextend}
\end{equation}
where $A_v = \mathcal{A}_v^T \mathcal{A}_v$, by definition.
We know the expression for $\frac{\partial \mathcal{L}}{\partial A_v}$ from Eq.~\ref{eqn:supp_vid}, but to obtain the expression for $\frac{\partial }{\partial \mathcal{A}_v}\mathcal{A}_v^T \mathcal{A}_v$ we define a permutation matrix $Q_{m,n}$ of size $mn \times mn$ (as in \cite{matcalc,gcrf2})  as follows:
\begin{equation}
 Q_{m,n} \text{vec}(M) = \text{vec}(M^T)\text{,}
\end{equation}
where vec$(M)$ is the vectorization operator that vectorizes a matrix $M$ by stacking its columns. Thus, the operator $Q_{m,n}$ is a permutation matrix, composed of $0$s and $1$s, and has a single $1$ in each row and column. 
When premultiplied with another matrix, $Q_{m,n}$  rearranges the ordering of rows of that matrix, while when postmultiplied with another matrix, $Q_{m,n}$ rearranges its columns.
Using this matrix, we can form the following expression \cite{matcalc}:
\begin{equation}
 \frac{\partial }{\partial \mathcal{A}_v}\mathcal{A}_v^T \mathcal{A}_v = \left({\bf I} \otimes \mathcal{A}_v^T\right) + \left(\mathcal{A}_v^T \otimes {\bf I}\right) Q_{D,N} \text{,}
 \label{eqn2:aat}
\end{equation}
where ${\bf I}$ is the $N\times N$ identity matrix. Substituting Eq.~\ref{eqn:supp_vid} and Eq.~\ref{eqn2:aat} into Eq.~\ref{eqn2:kernelextend}, we obtain:
\begin{equation}
 \frac{\partial \mathcal{L}}{\partial \mathcal{A}_v} = - \left(\frac{\partial \mathcal{L}}{\partial \mathbf{b}_v}  \otimes \mathbf{x}_v\right)\left( \left({\bf I} \otimes \mathcal{A}_v^\top\right) + \left(\mathcal{A}_v^\top \otimes {\bf I}\right) {Q_{D,N}}\right) \label{eqn2:grad_s}.
 \end{equation}

\subsection{Gradients of Temporal Embeddings}
As in the last section, from \refeq{eqn:supp_qi2}, we ignore any terms that do not depend on $\mathbf{b}_v$ or $T_{v,u}$ and write it as $\mathbf{b}_v = c + \sum_{u \ne v}  T_{v,u} \mathbf{x}_u $.

Using the strategies in the previous section and the sum rule of differentiation, the gradients of the temporal embeddings are given by the following form:
\begin{equation}
 \frac{\partial \mathcal{L}}{\partial \mathcal{T}_v} = - \sum_{u} \left(\frac{\partial \mathcal{L}}{\partial \mathbf{b}_u}  \otimes \mathbf{x}_v\right)\left( \left({\bf I} \otimes \mathcal{T}_u^\top\right) + \left(\mathcal{T}_u^\top \otimes {\bf I}\right) {Q_{D,N}}\right) \label{eqn2:grad_s}
 \end{equation}

{\small
\bibliographystyle{ieee}
\bibliography{egbib}
}

\end{document}